\documentclass[letterpaper]{article} 
\usepackage{aaai24}  
\usepackage{times}  
\usepackage{helvet}  
\usepackage{courier}  
\usepackage[hyphens]{url}  
\usepackage{graphicx} 
\usepackage{natbib}  
\usepackage{caption} 
\usepackage{algorithm}
\usepackage{algorithmic}

\usepackage{newfloat}
\usepackage{listings}
\DeclareCaptionStyle{ruled}{labelfont=normalfont,labelsep=colon,strut=off} 
\floatstyle{ruled}
\newfloat{listing}{tb}{lst}{}
\floatname{listing}{Listing}
\usepackage{amsmath}  
\usepackage{amsfonts}  
\usepackage{subcaption}		

\usepackage{todonotes}		

\newcommand{\vx}[1]{\mathbf{#1}}

\title{Harnessing Manycore Processors with Distributed Memory for Accelerated Training of Sparse and Recurrent Models}
\author{
    Jan Finkbeiner\textsuperscript{\rm 1, 2},
    Thomas Gmeinder\textsuperscript{\rm 3},
    Mark Pupilli\textsuperscript{\rm 3},
    Alexander Titterton\textsuperscript{\rm 3},
    Emre Neftci\textsuperscript{\rm 1, 2}
}
\affiliations{
    \textsuperscript{\rm 1}Research Center Juelich\\
    \textsuperscript{\rm 2}RWTH Aachen University\\
    \textsuperscript{\rm 3}Graphcore\\

    j.finkbeiner@fz-juelich.de, 
    thomasg@graphcore.ai,
    markp@graphcore.ai,
    alexandert@graphcore.ai,
    e.neftci@fz-juelich.de
}

\begin{document}

\maketitle

\begin{abstract}
Current AI training infrastructure is dominated by single instruction multiple data (SIMD) and systolic array architectures, such as Graphics Processing Units (GPUs) and Tensor Processing Units (TPUs), that excel at accelerating parallel workloads and dense vector matrix multiplications. Potentially more efficient neural network models utilizing sparsity and recurrence cannot leverage the full power of SIMD processor and are thus at a severe disadvantage compared to today's prominent parallel architectures like Transformers and CNNs, thereby hindering the path towards more sustainable AI. 
To overcome this limitation, we explore sparse and recurrent model training on a massively parallel multiple instruction multiple data (MIMD) architecture with distributed local memory. We implement a training routine based on backpropagation through time (BPTT) for the brain-inspired class of Spiking Neural Networks (SNNs) that feature binary sparse activations. We observe a massive advantage in using sparse activation tensors with a MIMD processor, the Intelligence Processing Unit (IPU) compared to GPUs. On training workloads, our results demonstrate 5-10$\times$ throughput gains compared to A100 GPUs and up to 38$\times$ gains for higher levels of activation sparsity, without a significant slowdown in training convergence or reduction in final model performance. Furthermore, our results show highly promising trends for both single and multi IPU configurations as we scale up to larger model sizes. Our work 
paves the way towards more efficient, non-standard models via AI training hardware beyond GPUs, and competitive large scale SNN models.
\end{abstract}
\section{Introduction}
The hardware lottery hypothesis \cite{hooker21a} contends that the current dominance of Graphics Processing Units (GPUs) in the AI hardware market constrains the innovation of model architectures.
As a result, contemporary research in Artificial Intelligence (AI) and Machine Learning (ML) is predominantly driven by parallel, dense architectures
that fully leverage the processing power of GPUs. 
However, this situation restricts the exploration of alternative approaches with great potential, such as neuromorphic computing, due to the inherent inefficiency of such architectures on GPUs.

In response to the limitations of traditional computing paradigms, neuromorphic computing has emerged as a promising alternative to achieve the efficiency of the human brain \cite{Mead90_neurelec,Chicca_etal13_neurelec,Benjamin_etal14_neurmixe,Friedmann_etal17_demohybr,Davies_etal18_loihneur,Frenkel_Indiveri22_reck28nm}. 
Distributed local memory, where processing and memory are closely intertwined is a key characteristic of brains, neuromorphic hardware and physical computing in general.
In the brain, synapses both store information and perform computation, resulting in no additional necessary data transfer as prevalent in today's von Neumann architectures. 
The other two core characteristics of the brain are recurrence and sparsity.
Recurrence enables the reuse of locally stored information and weights, thereby minimizing data transfer. 
Data transfer across processors is the most costly operation in modern technology \cite{Horowitz14_11}. 
Reducing transfer via sparsity can lead to significant energy efficiency and throughput gains.
Together, networks mimicking these brain-like characteristics can pave the way towards an energy-efficient and sustainable future for large-scale model deployment. 
In this article, we demonstrate an architecture emphasizing these characteristics, namely local distributed memory and a novel algorithm for training networks featuring sparsity and recurrence as in the brain that vastly outperforms their equivalent implementations in GPUs in throughput.
GPUs (and TPUs \cite{Jouppi_etal17_indaperf}), the prevalent hardware for deep learning, are massively parallel single instruction multiple data (SIMD) architectures, optimized for processing large chunks of data and executing identical operations in parallel. 
However, networks featuring recurrence and sparsity and thereby sequential computations, cannot fully utilize the parallel processing power of GPUs. Using a multiple instruction mutliple data (MIMD) manycore processor with distributed in-processor memory, we demonstrate speed-ups reaching 20$\times$ on training tasks for especially challenging network architectures featuring both recurrence and dynamic sparsity, namely spiking neural networks (SNNs).

SNNs are a class of neuromorphic algorithms that emulate key computational principles of biological neurons.
Unlike traditional Artificial Neural Networks (ANNs), which output continuous-valued activations and redundant computations, SNNs transmit information through discrete spike events, potentially resulting in significant energy savings \cite{Zenke_etal21_visujoin}. 
Their sensor counterpart, \emph{i.e.} event-based cameras, departs from fixed-time interval sampling, responding asynchronously to changes in stimuli, leading to reduced data redundancy and ultra-low data rates \cite{Gallego_etal19_evenvisi}.
This feature has proven ground-breaking for applications requiring high temporal resolution, such as robotics (control) and agile drone flight \cite{rosinol18a}. 
SNNs are ideally suited to process such event-based sensor data. 

Still, training large-scale SNN models and generally models featuring recurrence and sparsity remains prohibitively expensive due to the inadequacies of widely used training hardware.

To address these constraints and foster the exploration of alternative hardware architectures, this work focuses on a massively parallel multiple instruction multiple data (MIMD) architecture with distributed local memory, the Intelligent Processing Unit (IPU). 
By distributing neurons on dedicated tiles of the IPU and by implementing binary sparse representations for the SNN's sparse spike activation vector, we take advantage of the IPU's memory locality and efficient fine-grained memory access. This way we both minimize data transfer within and between IPUs and maximize the computational efficiency of the implementation. By examining the IPU's capabilities, we aim to unlock new avenues for AI and ML research, breaking free from the hardware lottery hypothesis and opening the door to innovative, efficient, and biologically plausible computing systems.
\subsection{Related Work}
Sparse neural networks have garnered substantial interest for their potential to enhance both training and inference efficiency. A multitude of research efforts have been directed towards developing algorithmic and hardware approaches to harness sparsity's benefits \cite{hoefler21a}. However, achieving substantial gains in throughput or training time on SIMD and systolic array based architectures like GPUs and TPUs remains challenging. Notably, the SIMD nature of GPUs, including the incorporation of Tensor Cores, presents limitations for efficiently accelerating sparse operations \cite{gale20a, huang22a}. Consequently, achieving benefits from sparse matrix matrix multiplication (SpMM) in the absence of specific sparsity patterns proves difficult, except for extreme sparsity levels of around 99$\%$ and higher \cite{hoefler21a}. To address this, techniques have emerged to introduce and exploit structure within sparse matrices \cite{huang22a, mishra21a, wang20a}. For instance, NVIDIA's 2:4 sparsity support optimally utilizes tensor cores \cite{mishra21a}. Additionally, strategies like the Inspector-Executor Framework and Autotuning \cite{wang20a} as well as other techniques to optimize data layouts for memory access and load balancing \cite{gale20a} have been employed.

While existing techniques predominantly focus on inference acceleration using static sparsity patterns in pruned models, dynamic sparsity patterns are also gaining attention. These encompass evolving weight sparsity during training or rapid changes in activation-based sparsity, particularly in event-driven recurrent architectures \cite{subramoney23a, knight22a, eshraghian21a}. However, only a few works effectively leverage sparse activations for throughput gains \cite{knight22a, perez21a}. Recurrent neural network architectures, such as LSTM, GRU, and SNNs, face challenges due to their recursive execution compared to inherently parallel models like Transformers. Specific efforts enable parallel timesteps execution via customized CUDA kernels while restricting SNN architectures to feed-forward structures \cite{shrestha18a, bauer23a, fang20a}.

For neuroscience-driven spiking neural network simulations with binary activations and no gradient computation, scalable GPU and CPU simulators have been developed \cite{yavuz23a, knight23a, niedermeier22a, golosio21a, gewaltig07a}.  Beyond conventional hardware, neuromorphic hardware aims to enable power efficient and fast execution of brain-inspired recurrent and sparse models, specifically SNNs. However, these hardware architecutres are mainly geared towards inference of SNN models or brain inspired learning rules that rely on local information. Prominent examples include TrueNorth\cite{Merolla_etal14_millspik}, Loihi \cite{Davies_etal18_loihneur}, Spinnaker \cite{Furber_etal14_spinproj}, and the mixed-signal Brainscales \cite{Friedmann_etal17_demohybr} architecture. Several FPGA implementations have been realized to accelerate both biological SNN simulations \cite{Kauth_etal23_neurdesi,Wang_etal18_fpgamass} as well as CNN applications with sparse activations \cite{aimar19a}.

Prior exploration of SNN behavior on the IPU has produced inconclusive results in comparison to GPU or TPU implementations: Biologically inspired neuron dynamics implementations on various accelerators reveal non-favorable IPU performance \cite{landsmeer23a}. Another study \cite{sun22a} demonstrates enhanced throughput for single hidden layer architectures, but its results for more complex designs remain unclear. Notably, these studies do not optimize code beyond IPU compilation, neglecting the potential of tile-locality and sparse activation tensors for given SNN models, and were limited in scale. In contrast, our work focuses on exploiting the unique IPU architecture to take advantage of recurrence dynamic activation sparsity. Thereby we capitalize on both tile-locality and sparse activation tensors, enhancing performance via a dedicated sparse matrix multiplication algorithm. Additionally, this renders our implementation more memory efficient allowing the deployment of larger scale models.

Summarizing, our contributions are the following: 
\begin{itemize}
    \item As a model for sparse and recurrent architectures, we implement a sparse SNN training algorithm based on BPTT on a manycore processor, the IPU. 
    \item We explicitly capitalize on the hardware's in-processor memory by distributing neurons in a tile-local fashion leading to acceleration factors of up to 20$\times$ compared to GPU implementations.
    \item We verify that our sparse training algorithm maintains final accuracies and convergence rates on-par with the dense equivalent.
    \item We demonstrate the scalability of our implementation to larger model sizes and mulit-IPU implementations.
\end{itemize}
\section{Methods}
\subsection{The Graphcore IPU}
The Intelligence Processing Unit (IPU) \cite{jia19a} by Graphcore is a massively parallel compute architecture with distributed local memory ideally suited for multiple data multiple instruction (MIMD) workloads. 

The IPU programming paradigm based on the Bulk Synchronous Parallel (BSP) model features the sequential execution of multiple supersteps composed of a local computation phase, a communication phase, and a barrier synchronization phase.
During each computation phase, every core has access to 624\,kB of its dedicated SRAM. The core and its dedicated SRAM together form a \emph{tile}.

In this work, the IPU is programmed via the C++ based Poplar SDK for fine-grained control over the computational graph and memory allocation. 
Additionally a Python API via all major machine learning libraries (Pytorch \cite{pytorch2019}, Tensorflow \cite{tensorflow2015whitepaper}, ONNX, Jax \cite{jax2018github}) gives simple and flexible access to the IPU. 

The IPU features low latency, high bandwidth intra- and inter-IPU interconnect. 
This feature becomes especially useful when designing large scale, distributed networks. 

The IPU architecture has the potential to provide improvements over GPU and TPU architectures for recurrent neural network architectures with sparse connectivity and activations.
This is because sequential memory read and write operations between computation operations are faster on the IPU as memory latency is extremely low (equivalent to L1 caches on GPUs). 
Furthermore, the IPU can efficiently utilize unstructured sparsity compared to GPUs, as it allows for efficient reading and writing of small data packets (8 bytes for highest bandwidth) compared to the minimal data read size of 128 bytes (32 values of 32 bits) for NVIDIA GPUs. 
Therefore we believe it to be especially well suited for SNNs which combine both recurrence and sparsity. 

\subsection{Spiking Neural Network (SNN) Training}\label{sec:background:snns}
SNNs can be viewed as special cases of recurrent neural networks with binary and sparse activations and a specific set of internal dynamics \cite{Neftci_etal19_surrgrad}.
Additionally, complex connectivity structures like explicit recurrence where higher levels feed back to lower layers are more common in brain-inspired networks and SNN architectures \cite{Kubilius_etal19_braiobje} than for standard RNN architectures, like GRUs or LSTMs, which are typically implemented as feed-forward architectures with local recurrence.

The spiking neuron dynamics can be described by various sets of differential equations.
In this work we chose the standard leaky-integrate-and fire (LIF) neuron model, though our approach can be easily generalized to more complex neuron dynamics. 
Equations (\ref{eq:lif:I}-\ref{eq:lif:S}) below show the time-discretized versions of the LIF equations that are implemented in our models:
\begin{align}
I_i[t+1] &= f(I_i[t], \vx{S}^\mathrm{in}[t]) = \sum w_{ij} \; S_j^\mathrm{in}[t], \label{eq:lif:I} \\
u_i[t+1] &= \alpha \; u_i[t] ( 1 - S_i^\mathrm{out}[t] ) + (1-\alpha)  \; \frac{1}{C} \; I_i[t] \label{eq:lif:u}, \\
S_i^\mathrm{out}[t] &= \Theta (u_i[t] - \vartheta_i) . \label{eq:lif:S}
\end{align}
Equation (\ref{eq:lif:I}) describes the calculation of the input current $I_i$ of neuron $i$. Generally, $I_i$ can be formulated as a function of the current at the previous timestep and the spikes. 
The synaptic weights $w_{ij}$ are the adjustable and trainable parameters. 
Equation (\ref{eq:lif:u}) describes the dynamics of the LIF-neuron's membrane potential $u_i$ including a reset mechanism that resets the membrane potential to zero after a spike $S^{\text{out}}_i$ was emitted. 

Equation (\ref{eq:lif:S}) defines the spike generation function modeled via the Heaviside function $\Theta$. 
The objective function can be minimized with respect to the synaptic weights by using stochastic gradient descent (SGD) generally result in good model performance when training SNNs \cite{Neftci_etal19_surrgrad}. 
The calculation of gradients across the non-differentiable spiking activation function \ref{eq:lif:S} was enabled by the surrogate gradient approach \cite{Zenke_Ganguli18_supesupe,Neftci_etal19_surrgrad}. The surrogate gradient approach introduces a separate function to overwrite the spiking activations behavior in the backward pass for the gradient computation. However, it does not alter the behavior in forward pass. Different choices for such surrogate functions exist \cite{Neftci_etal19_surrgrad} with mostly similar performance \cite{Zenke_Vogels20_remarobu}. Here, we choose the SuperSpike surrogate function \cite{Zenke_Ganguli18_supesupe} as defined in Equation \ref{eq:superspike}.
\begin{align}
S(x) &= \Theta ( x ) , \\
\frac{\partial S(x)}{\partial x} &= h(x) = \frac{1}{\left( \beta |x| + 1 \right)^2} . \label{eq:superspike}
\end{align}
Using backpropagation through time (BPTT) the gradients of the loss with respect to the weights $\frac{\partial L}{\partial w_{ij}}$ can be calculated iteratively backwards though time, which enables the use of standard neural network optimizers. In this work either Adam \cite{Kingma_Ba14_adammeth} or plain stochastic gradient descent (SGD) is used. 
\begin{figure*}[t]
  \includegraphics[width=\linewidth]{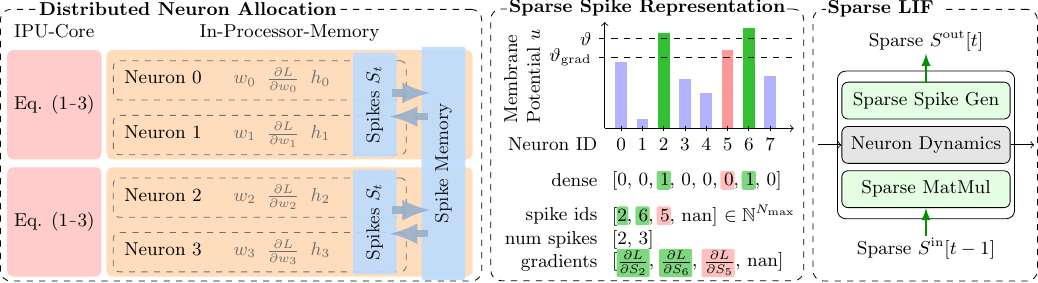}
  \caption{SNN implementation on the Graphcore IPU. Left: Illustration of the neuron placement on dedicated IPU-tiles including weights $w$, gradients $\frac{\partial L}{\partial w}$ and states $h$, which minimizes the necessary data transfer to solely spike based input and output data. Middle: Dense vs sparse spike vector representation as generated from membrane potentials and threshold vectors. Right: Computational graph of our sparse LIF cell with operations involving sparse spiking vectors marked in green.}
  \label{fig:snn_ipu_implementation}
\end{figure*}
\subsection{Exploiting SNN Sparsity on the IPU} \label{sec:implementation}
We implemented a BPTT based training routine for multi-layer spiking neural networks (SNNs) following Equations (\ref{eq:lif:I}-\ref{eq:lif:S}) on the Graphcore IPU and use sparse spiking vectors together with a sparse matrix multiply algorithm that utilizes the IPU's MIMD nature and ability of fine-grained low latency memory access. The IPU's design featuring distributed in-processor memory with parallel processes fits well with the brain's principle of distributed local computation and therefore with the deployment of SNNs in neuromorphic hardware.

\subsubsection{Distributed Neuron Allocation} As illustrated in Figure \ref{fig:snn_ipu_implementation}, the IPU's distributed in-processor memory allows allocating neurons on a dedicated tile for the whole training process, which includes the simulation of their dynamics, the synaptic weights and weight gradients and the input current calculation. Only the spike tensors and auxiliary data during their computation have to be communicated between tiles. This drastically reduces data transfer on the chip compared to other architectures which rely on external memory, like GPUs or TPUs, and thereby improves efficiency.
\subsubsection{Sparsifying Forward and Backward pass} The right panel in Figure \ref{fig:snn_ipu_implementation} shows the computational graph of the forward pass of a single layer of spiking neurons, featuring a function that computes the input current based on input spikes (Eq. \ref{eq:lif:I}), a function that integrates the neuron dynamics based on the input current (Eq. \ref{eq:lif:u}) and a function that generates the output spikes based on the membrane potential (Eq. \ref{eq:lif:S}). The integration of internal neuron dynamics is naturally modeled as a dense function where every neuron's state variables are updated every timestep. While the spiking activations and the input generating functions are typically also modeled as dense tensors and dense functions, in this work we utilize the sparsity of these variables to obtain efficiency gains with improved throughput and latency. For linear layers the forward pass is simplified from a matrix multiplication in the dense case, to a simple read-and-sum operation in the sparse case. Similarly, the corresponding functions in the backward pass can benefit from the generated sparse spiking tensor representations.
\subsubsection{Sparse Spike representation} In order to make use of the sparse activation of SNNs we use a sparse spike representation that stores the indices and the number of neurons that spiked. Our choice of sparse spike representation is driven by two factors:
\begin{itemize}
\item A naive implementation that contains only information about neurons that spiked can lead to inaccurate gradients in the backward pass. We circumvent this issue by taking a very similar approach as in \cite{perez21a}: In addition to propagating spikes from neurons whose membrane potential has crossed the spiking threshold $\vartheta$, information is also transmitted from neurons above a secondary threshold $\vartheta^\mathrm{grad}$.
\item The IPU is constrained to static tensor sizes that must be known during compile time. It is not possible to allocate or communicate varying tensor sizes depending on the number of spikes. Therefore we define the maximal number of considered spikes $N_\mathrm{max}$ as a hyperparameter and allocate spike tensors accordingly. This requires allocating a second tensor that contains the number of spikes.
\end{itemize}
Based on these two considerations we define the sparse spike representation for a given batchsize $B$ as illustrated in the middle panel of Figure \ref{fig:snn_ipu_implementation}: It contains spike ids $\in \mathbb{N}^{B \times N_\mathrm{max}}$, the  number of spikes and gradients $ \in \mathbb{N}^{B \times 2}$ and the spike gradient tensor $ \in \mathbb{R}^{B \times N_\mathrm{max}}$, which is only present in the backward pass. If the number of neurons above threshold exceeds $N_\text{max}$ at a certain timestep, excess spikes are randomly dropped.
\subsubsection{Integration into the SNN training workflow} The current work features two sparse implementations for the IPU: Firstly, we developed a custom fully connected multi-layer LIF network in order to achieve maximal gains in regard to throughput and memory reduction. 
Secondly, we enable the integration of custom dynamic sparse operations into a standard implementation of neuron dynamics with Tensorflow's Python API. This gives flexibility to the programmer in the choice of neuron model and network architecture while still taking advantage of sparse activations. The first, fully custom approach is used to generate the benchmarking results.
\section{Results}
\begin{figure*}[t]
\centering
\includegraphics{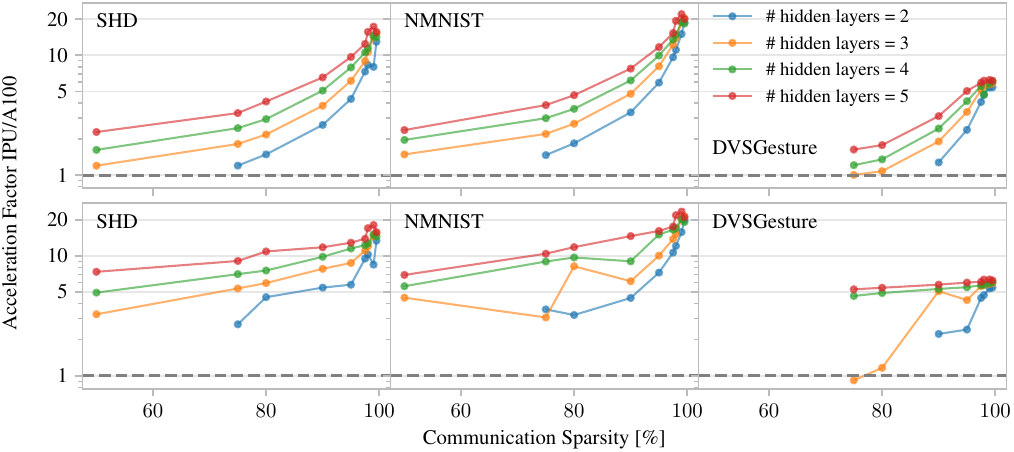}
\caption{Acceleration factor of sparse IPU vs. dense A100 GPU implementations across varying levels of communication sparsity (1 - maximal activity) for a fixed network size of 2944 neurons with different hidden layer configurations on the SHD (left), NMNIST (middle) and DVSGesture (right) datasets. Top: ``fixed activity'' mode, all neurons spike every timestep, representing a lower bound approximation of throughput. Bottom: ``natural activity'' mode, where actual sparsity can exceed communication sparsity set by $N_\mathrm{max}$ resulting in maintained high acceleration factors for lower levels of communication sparsity.} 
\label{fig:results:sparsity}
\end{figure*}
\begin{figure*}[t]
\centering
\includegraphics{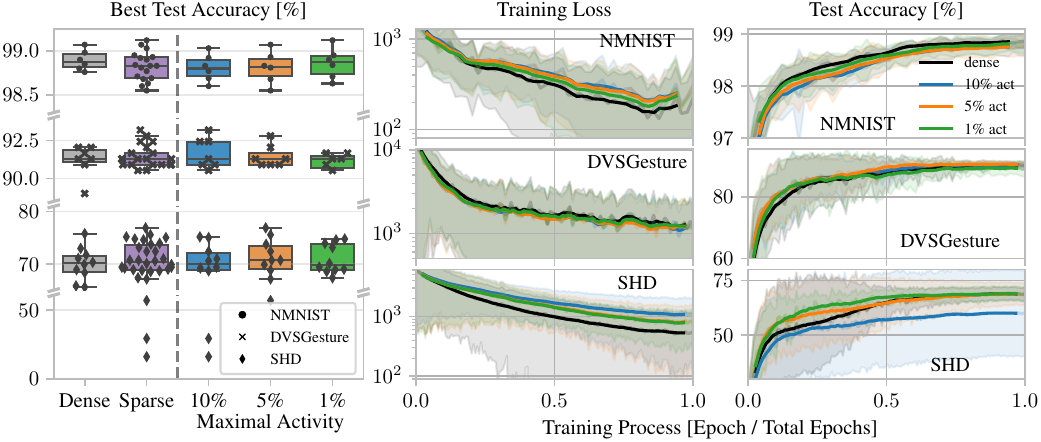}
\caption{Achieved test accuracy (left) and convergence of training loss (middle) and test accuracy (right) of dense and sparse implementations for the same set of hyperparameters on NMNIST (top), DVSGesture (middle) and SHD (bottom) dataset. Slight differences in training loss convergence between dense and sparse implementations show no impact on test accuracy.\vspace*{-0.2cm}}
\label{fig:results:accuracy}
\end{figure*}
\begin{figure*}[t]
\centering
\includegraphics{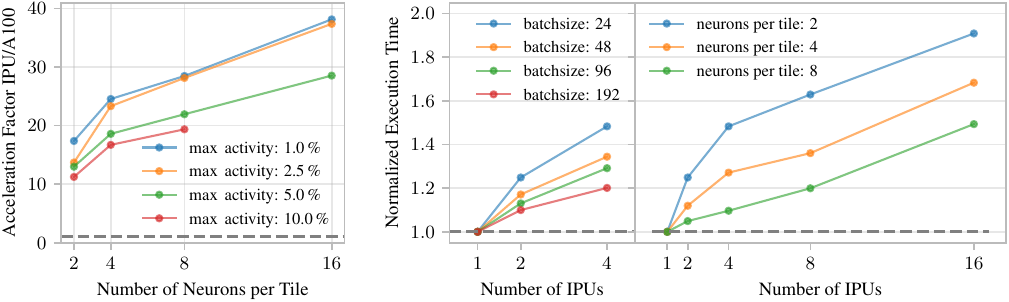}
\caption{Left: Acceleration factor of Sparse IPU vs. A100 implementation for diverse network sizes achieved by adjusting number of neurons per IPU-tile. Right: Weak scaling results on SHD dataset with ``natural activity'' at max activity of 5$\%$, exploring compute-workload to communication-latency ratios through batch size variation and network size per IPU adjustment.}
\label{fig:results:scaling}
\end{figure*}
\subsection{Benchmarking Setup} \label{sec:bench_setup}
We benchmarked the sparse SNN implementation for the IPU by evaluating the throughput of multiple SNN models using the SHD \cite{Cramer_etal20_heidspik}, N-MNIST \cite{Orchard_etal15_realeven} and DVSGesture \cite{Amir_etal17_lowpowe} dataset and compared to an equivalent dense implementation on a GPU (NVIDIA GeForce RTX 3090, NVIDIA V100, NVIDIA A100). We demonstrate this for fully-connected multi-layer feed-forward SNN architectures. If not specified otherwise, the network contains 2 neurons per tile, totaling $2 \times 1472=2944$ neurons. While the input and output sizes are dataset dependent, the remaining neurons are equally distributed into the specified number of hidden layers. The implementation is based on Keras \cite{chollet2015keras} with Tensorflow \cite{tensorflow2015whitepaper} backend on the Python side and custom code via the Poplar SDK for the IPU in C++. 
The execution time is measured via a python timer within a Keras callback. 
All implementations use 32-bit floating point precision, except for integer valued sparse spike tensors. The dense GPU implementation utilizes tensor cores using the \texttt{tf.fp32} datatype. To further improve throughput we choose a large value for \texttt{steps\_per\_execution} to process the whole data either within 1 or 2 cycles and we set \texttt{jit\_compile=True} for the GPU implementation. 
We discard the first obtained time per epoch due to possible initial slowdows as a sideffect of compilation times and other overheads and average over the remaining times. 
We load the data to CPU RAM before execution to prevent a degradation in throughput due to bottlenecks in reading from disk. 
We fix the batch size for all benchmarking experiments to 48 to reduce the number of hyperparameters to sweep. While we observe even greater gains on the IPU for smaller batchsizes, we believe a batchsize of 48 to be a fair and insightful size for real world usecases.
In order to  obtain reliable results that are independent of random weight initialization and other factors, we execute the runs on the IPU with the sparse algorithm implementation with two different settings:
\begin{enumerate}
\item Natural activity: We don't manipulate the spiking activity, but randomly initialize the weight matrices and let the network run. For deeper layers, the activity tends to converge towards zero. This can be viewed as a conservative approximation of the \textbf{upper bound in throughput} and acceleration compared to the GPU runs.
\item Fixed activity: We force all neurons to spike. This leads to a saturation in the spike vectors, and thereby fixes the activity to the maximally allowed activity, which is chosen via $N_\mathrm{max}$ for the spike vector sizes. This mode can be viewed as approaching the \textbf{lower bound in throughput}.
\end{enumerate}
The input spike tensors are fixed to a dataset dependent size, that was chosen based on the activity distribution of the data (NMNIST: 32, SHD: 48, DVSGesture: 96).
The IPU and benchmarking code is available at https://github.com/PGI15/SparseSNNsIPU.

For easy of use and reproducibility, the experiments for training convergence and accuracy are performed using a sparse tensor implementation in Jax with custom CUDA kernels for execution on the GPU. The custom GPU code replicates our custom implementation on the IPU with sparse tensors and sparse matrix multiplication. The code is available at https://github.com/PGI15/SparseSpikesJax.
\subsection{Experimental Evaluation}
Our results show a clear advantage in using sparse activations with our multi-layer SNN implementation on the IPU, demonstrating increasing gains with increasing sparsity of the activations. We do not observe significant  differences in performance across different NVIDIA GPUs (RTX 3090, V100, A100) leading us to focus on results relative to the NVIDIA A100 in our discussions. The acceleration factor is determined as the ratio of GPU batch processing time to IPU batch processing time: $\mathrm{acceleration~factor} = \frac{\mathrm{time~per~batch~on~the~GPU}}{\mathrm{time~per~batch~on~the~IPU}}$.

\subsubsection{Acceleration on Single IPU} Figure \ref{fig:results:sparsity} shows the acceleration factor on the IPU, where Figure \ref{fig:results:sparsity} (top) demonstrates the behavior for the ``fixed activity'' and Figure \ref{fig:results:sparsity} (bottom) for the ``natural activity'' case. The measurement of the acceleration factor includes all necessary operations for the training process, meaning the forward pass, the backward pass to calculate the gradients, and the weight update. By utilizing the sparse activations we achieve substantial gains in throughput on the IPU by at least a factor of 5-10$\times$ compared to the GPU. For more extreme levels of activation sparsity which are still relevant for training runs in practice, even higher acceleration factors of 15-20$\times$ can be achieved. Averaging over all runs up to including maximal activity of $10\,\%$, resulting in an average communication sparsity of $97.8\,\%$, we obtain a throughput of 1.472 million event-frames or 14.72 thousand sequences (100 timesteps) per second for the NMNIST dataset using the sparse implementation on the IPU. The dense implantation on the A100 achieves 102 thousand frames or 1.02 thousand sequences per second (batchsize 48, architecture: [2048, 2934 neurons split over $n$ layers, 10]).
Somewhat reduced gains are observed for the DVSGesture dataset. Here, the network architecture showcases less homogeneity, particularly with a considerably larger input layer, revealing the necessity for enhancements in our mapping strategy to achieve superior load balancing on the IPU. 
Comparing ``fixed activity'' and ``natural activity'' scenarios, we observe that the former shows the anticipated proportional scaling with activity, while the latter maintains high acceleration factors for lower sparsity values. This suggests that execution speed is more influenced by network activity rather than specific hyperparameters limiting maximal activity. This fact underscores the robustness of our IPU implementation in the face of varying sparsity hyperparameters that might overestimate network activity, thus minimally compromising execution speed in practical applications.

\subsubsection{Training Convergence and Accuracy} Our analysis on final training accuracy and training convergence (Fig. \ref{fig:results:accuracy})  aligns with prior research \cite{perez21a}, indicating that sparse activation tensors don't impede training convergence or final model performance. Thus, the observed acceleration factors in throughput (Fig. \ref{fig:results:sparsity}) directly translate to faster overall training times. In order to fairly assess the differences in training behaviour, we initially determine the best performing set of hyperparameters (learning rate, number of time steps, decay constants, ...) for the dense implementation. Subsequently, other hyperparameters, namely the number of hidden layers and the use of augmentations, were varied for both the dense and the sparse implementations leading to the showcased training runs in our analysis. The left panel in Figure \ref{fig:results:accuracy} shows the achieved test accuracy on the NMNIST, DVSGesture and SHD dataset across dense and sparse implementations, factoring in varying levels of sparsity. Except for a few outliers that did not converge in the sparse case, the test accuracies for dense and sparse implementations exhibit similar distributions. Notably, the best result on all three datasets was achieved with a sparse implementation. While some slowdowns occur in training loss convergence (Figure \ref{fig:results:accuracy} middle), the corresponding test accuracies (Figure \ref{fig:results:accuracy} right) show neither a reduced convergence speed nor a reduction in final accuracy. These results suggest an implicit regularization effect of the sparse activation tensors similar to Dropout \cite{Hinton_etal12_imprneur}. 

\subsubsection{Scalability Analysis} In order to analyse the scalability of our implementation, we both deploy larger networks on a single IPU by placing more neurons on every tile, and we perform weak scaling experiments, where the network size per IPU stays constant but is replicated to multiple IPUs. 

Figure \ref{fig:results:scaling} (left) shows that for increasing network size on a single IPU, the acceleration compared to the GPU baseline further increases. To increase the network size, we modify the number of neurons that are allocated on each tile and extend the number of layers in the network architecture accordingly. Due to memory constraints we reduced the number of timesteps to 10 for the displayed experiments and used SGD instead of Adam. While these configurations may not align with standard training setups, they nevertheless provide informative insights and should translate to practical setups. Collectively, we observe that the acceleration is more pronounced in the case of the ``natural activity'' but also present for the ``fixed activity'' setup and it differs across datasets. 
The advantageous scaling on the IPU provides evidence that scaling up the network size on a single IPU in production settings further benefits acceleration compared to the GPU. 

Figure \ref{fig:results:scaling} (right) shows the results of our weak scaling experiment for the SHD dataset. Here, the number of IPUs is increased while simultaneously increasing the network size (in terms of number of layers) accordingly. The number of neurons per tile and therefore the workload per tile and IPU is thereby kept constant. The ideal scaling behavior would lead to no slowdown in the total execution time for multi IPU settings as indicated by the grey dashed line. We analyse how the slowdown for multi-IPU settings behaves for settings ranging from 1 to 16 IPUs.
Primarily, we observe that the slowdown can be reduced by increasing the computational workload compared to communication latency. We demonstrate this by increasing the batch size (Figure \ref{fig:results:scaling}B left) and the number of neurons per tile meaning the network size per IPU (Figure \ref{fig:results:scaling}B right). We clearly observe that for increased network density per IPU we achieve better scaling results, lowering the slowdown from 1.9$\times$ to 1.5$\times$ for the shown experiment when executing on 16 IPUs compared to a single IPU run. Additionally, we observe very little slowdown for large batchsizes, where we observe minimal slowdowns of 1.2$\times$ for 4 IPUs for large batchsizes of 192 samples per batch. 
Overall, this demonstrates improved scalability with network and data size. 

In conclusion, exploiting sparse activations of SNNs and the IPU's MIMD nature with local, in-processor memory, significantly accelerates SNN training compared to modern GPUs. Positive scaling for increased network size on a single IPU and across multiple IPUs showcases the promise of accelerated large-scale SNN training on the Graphcore IPU.
\section{Discussion}
In this work, we have demonstrated the Graphcore IPU's potential for accelerating the training of spiking neural networks through the use of sparse spike vectors and a sparse vector matrix multiply algorithm. Specifically, we have implemented the backpropagation through time (BPTT) training algorithm and achieved a substantial increase in throughput for realistic training scenarios - at least 5-10$\times$ higher compared to traditional architectures, and even higher in some special but realistic cases.
Since the sparse training algorithm is not exact in the general case, one could suspect that the true acceleration factor in terms of model convergence may not be realized. 
However, we observed no slowdown in training convergence as a result of the introduced sparsity. This means that the acceleration in throughput directly translates to faster total training time. 
Looking ahead, we have noted promising scaling behavior for larger networks per IPU and in multi-IPU settings. Specifically, we have observed a significant increase in acceleration compared to GPUs when training larger networks per IPU, with up to 30-40$\times$ improvement for 16 neurons per tile. Our weak scaling results also indicate good scaling behavior, with only a 1.4$\times$  slowdown in throughput when scaling up to 16 IPUs, resulting in an effective 11$\times$ increase in compute.

As with GPUs, the amount of memory required to perform training using BPTT becomes a limiting factor for IPUs. This challenge is further amplified on IPUs due to the reduced total SRAM memory on the chip. Consequently, it becomes difficult to scale up to larger networks, where we observe even greater acceleration factors. Therefore, techniques like gradient checkpointing and exploring new training algorithms beyond BPTT such as three-factor rules and eligibility propagation \cite{Neftci_etal19_surrgrad,Bellec_etal20_soluto} is crucial to enable efficient implementation of spiking neural networks on hardware, particularly for problems and datasets that necessitate larger networks.

In combining such techniques with the observed acceleration factors on the IPU, we see a clear path towards large scale SNNs on the IPU and therefore the potential to revolutionize the deep learning model and hardware landscape.
\section{Acknowledgements}
This research was supported by NeuroSys as part of the initiative ‘Clusters4Future’ is funded by the Federal Ministry of Education and Research BMBF (03ZU1106XX). The authors gratefully acknowledge computing time on the supercomputer JURECA \cite{jureca2021} at Forschungszentrum Jülich, providing access to NVIDIA A100 GPUs and Graphcore IPUs (as part of the JURECA DC Evaluation Platform) studied in this work.

\bibliography{biblio_unique_alt,aaai24}

\section{Supplementary Information}

This document contains supplementary information for the paper with the title ``Harnessing Manycore Processors with Distributed Memory for Accelerated Training of Sparse and Recurrent Models''. Specifically, it gives more details on the implementation, the used network architecture sizes and provides additional data.

\subsection{Implementation Details}

\subsubsection{Initialization and Tensor allocation} \label{sec:impl:init_tensor_alloc}

Calling our custom Tensorflow operation to execute a multi layer SNN on the IPU leads to the allocation of several tensors. First, depending on the network size and architecture, it is determined how to distribute neurons onto the tiles of the given number of IPUs. We tested our implementation with up to 16 IPUs. 

Following, the weight $w_{ij}$ and state $h_i$ tensors are allocated. The state and weight tensors are distributed according to the determined neuron mapping. Weights are allocated on the same tile as the post-synaptic neurons. Additionally, the sparse spike tensor are allocated and mapped linearly over the IPUs. The spike tensor of a specific layer is allocated on the same IPU, but not necessarily on the identical tiles as the layer's neurons.

When the backward pass is invoked, similarly the tensors for weight gradients $\frac{\partial L}{\partial w_{ij}}$, state gradients $\frac{\partial L}{\partial h_i}$ as well as the spike gradient tensors $\frac{\partial L}{\partial S_i}$ are allocated. The weight and state gradient tensors mirror the neuron allocation, whereas the spike gradient tensors are mapped linearly.

The weight and state tensors as well as their respective gradient tensors stay allocated and remain with the fixed tile mapping as they were initialized in. The values do not have to be reallocated on other tiles for the whole training process.

\subsubsection{Forward and Backward pass} \label{sec:impl:fwd_bwd_pass}
The computational graph of SNN training can be split into iterative steps that process one batch. The iterative steps to process a single batch of data are
\begin{itemize}
\item Forward pass, to calculate states and activations and the loss
\item Backward pass, to calculate gradient of the loss with respect to the weights
\item Optimizer step, to update the optimizer state and calculate and apply the weight update.
\end{itemize}
While we use the already implemented optimizers that come with Tensorflow on the IPU, we implement custom forward and backward passes. 

The computational graph of the forward pass of a single layer of spiking neurons can be described by three main functions: 
\begin{enumerate}
\item A function that calculates the input current based on input spikes. This is typically implemented as a linear or convolution layer. 
\item A function integrates the neuron dynamics based on the input current. 
\item A spike generation function generating the output spikes based on the membrane potential, which is typically implemented as a Heaviside function and outputs dense spiking vectors containing ones and zeros. 
\end{enumerate}
In our case, the linear layers and the LIF neuron dynamics are mathematically described in Eq. (1-3) in the main text.
The neuron dynamics, function 2, are naturally modeled as dense functions, where every timestep the membrane potential of every neuron is updated. The other two functions, function 1 and function 3, operate with spiking tensors, either as input or as output. While they are typically also modeled as dense tensors and dense functions, in this work we utilize the sparsity of these variables to obtain efficiency gains with improved throughput and latency. For this, we model the spike vectors as sparse tensors as described in the section ``Sparse Spike representation'' and implement sparse algorithms.

For the backward pass the calculation of the weight gradients is drastically sparsified. The calculation of the input gradients is also more efficient as a sparse implementation, but it requires a less efficient reduction operation over multiple tiles.

The calculation of the output spikes in the forward pass, function 3, is more complex in the sparse case compared to the dense case.
It is implemented by locally identifying the above threshold neurons and then iteratively combining the sparse spike vectors over multiple tiles, until the sparse spike vector for a whole layer is constructed. 
The gradient operation, however, just as the other sparsified operations, can again benefit from the sparsity by only calculating and updating the gradients of the membrane potentials of the relevant neurons.

\subsection{Benchmarking Setup}

\begin{table*}[]
\centering
    \caption{Dataset dependent sizes used for all experiments. The only exception is singlue IPU scaling for the NMNIST dataset, where the input was further downscaled to \texttt{[24,24,2]} due to memory bottlenecks.}
    \label{tab:dataset_sizes}
\begin{tabular}{|c|c|c|c|c|c|}
\hline
\textbf{Dataset} & \textbf{Original Sensor Size} & \textbf{Downscaled Sensor size} & \texttt{INP\_SIZE} & \texttt{SPRASE\_INP\_SIZE} & \texttt{NUM\_CLASSES} \\
\hline
\hline
SHD & \texttt{[700, 1, 1]} & \texttt{[700, 1, 1]} & 700 & 48 & 20 \\ \hline
NMNIST & \texttt{[32, 32, 2]} & \texttt{[32, 32, 2]} & 2048 & 32 & 10\\ \hline
DVSGesture & \texttt{[128, 128, 2]} & \texttt{[48, 48, 2]} & 4608 & 96 & 11 \\ \hline
\end{tabular}
\end{table*}

To generate benchmarking results Tensorflow 2.11 and CUDA-11.7 was used. 
The training convergence runs were executed using Jax 0.4.7 and CUDA 12.1.
On the IPU side we used Poplar 2.6 and the corresponding provided tensorflow wheel for the IPU. 

Table \ref{tab:dataset_sizes} shows the dataset dependent input and output sizes in more detail.

The sparse activation tensor sizes of hidden layers, \texttt{HIDDEN\_LAYER\_SPARSE\_SIZES}, were calculated from the dense layers sizes of the hidden layers, \texttt{HIDDEN\_LAYER\_DENSE\_SIZES}, meaning the number of neurons in a layer, via the following python line:

\texttt{HIDDEN\_LAYER\_SPARSE\_SIZES = [int((MAX\_ACTIVITY*hid\_dense\_size//2)*2) for hid\_dense\_size in HIDDEN\_LAYER\_DENSE\_SIZES]}

which guarantees that spike vectors have an even size and a size of at least 2.


For details no provided in this document we refer to the additionally provided code.

\subsection{Acceleration factor over Sparsity}

\begin{figure*}[h]
\centering
\includegraphics{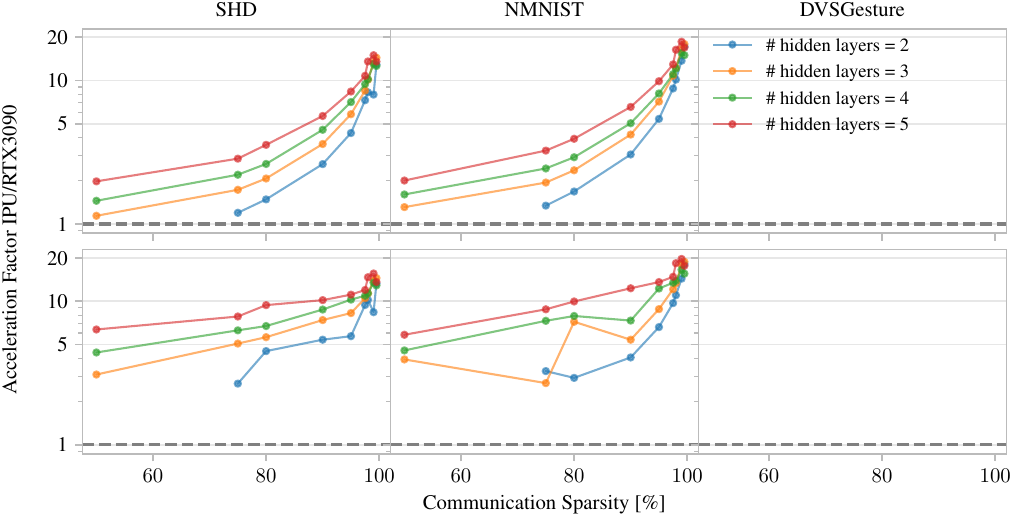}
\caption{Acceleration Factor of Sparse IPU vs. RTX3090 GPU implementations across varying levels of communication sparsity (1 - maximal activity) for a fixed network size of 2942 neurons with different hidden layer configurations on the SHD (left), NMNIST (middle) and DVSGesture (right) datasets. Top row: ``fixed activity'' mode, where all neurons spike every timestep, representing a lower bound approximation of throughput. Bottom row: ``natural activity'' mode, where actual sparsity can exceed the communication sparsity set by a fixed hyperparameter resulting in maintained high acceleration factors for lower levels of communication sparsity.} 
\label{fig:results:sparsity_rtx3090}
\end{figure*}

\begin{figure*}[h]
\centering
\includegraphics{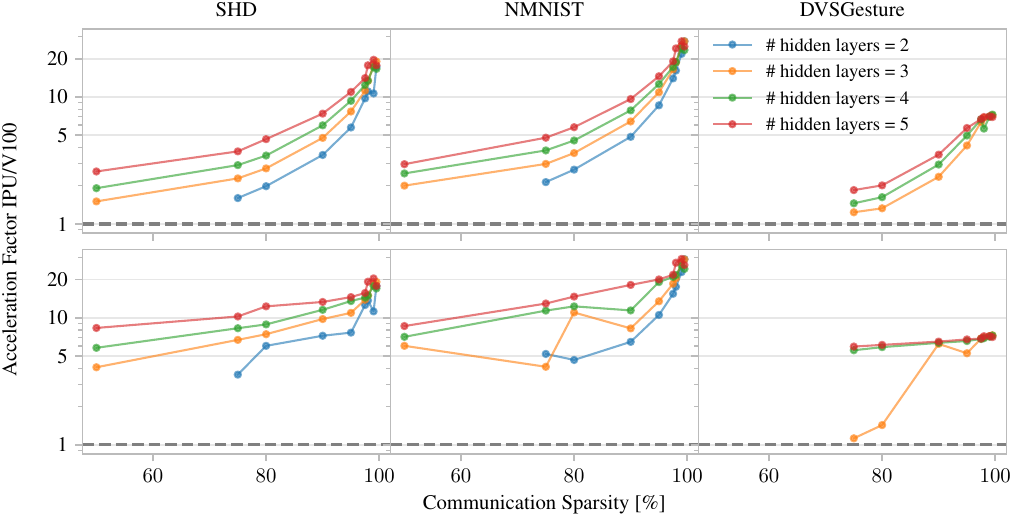}
\caption{Acceleration Factor of Sparse IPU vs. V100 GPU implementations across varying levels of communication sparsity (1 - maximal activity) for a fixed network size of 2942 neurons with different hidden layer configurations on the SHD (left), NMNIST (middle) and DVSGesture (right) datasets. Top row: ``fixed activity'' mode, where all neurons spike every timestep, representing a lower bound approximation of throughput. Bottom row: ``natural activity'' mode, where actual sparsity can exceed the communication sparsity set by a fixed hyperparameter resulting in maintained high acceleration factors for lower levels of communication sparsity.} 
\label{fig:results:sparsity_V100}
\end{figure*}

The specific network architecure sizes are described in Table \ref{tab:sparsity_network_architecure}. Figure \ref{fig:results:sparsity_rtx3090} and Figure \ref{fig:results:sparsity_V100} show the results against the NVIDIA RTX 3090 and V100 GPUs respectively.

\begin{table*}[]
\centering
    \caption{Network architecture single IPU scaling over sparsity experiments with a fixed number of 2 neurons per tile.}
    \label{tab:sparsity_network_architecure}
\begin{tabular}{|c|c|l|}
\hline
\# Hidden Layers & Total Number of Neurons & \texttt{DENSE\_SIZES} \\
\hline
\hline
2 & 2942 & [2312, 1466, 1466, 10] \\ \hline
3 & 2942 & [2312, 978, 978, 976, 10] \\ \hline
4 & 2942 & [2312, 734, 734, 732, 732, 10]  \\ \hline
5 & 2942 & [2312, 588, 586, 586, 586, 586, 10] \\ \hline
\end{tabular}
\end{table*}

\begin{table*}[]
\centering
    \caption{Network architecture used for training convergence and accuracy experiments. The variables \texttt{INP\_SIZE} and \texttt{NUM\_CLASSES} are dataset dependent as sepcified in \ref{tab:dataset_sizes}. }
    \label{tab:convergence_network_architecure}
\begin{tabular}{|c|l|}
\hline
\textbf{\# Hidden Layers} & \textbf{Layer Sizes} \\
\hline
\hline
2 & \texttt{[INP\_SIZE, 512, 128, NUM\_CLASSES]} \\
\hline
3 & \texttt{[INP\_SIZE, 1024, 512, 128, NUM\_CLASSES]} \\
\hline
4 & \texttt{[INP\_SIZE, 1472, 1024, 512, 128, NUM\_CLASSES]} \\
\hline
5 & \texttt{[INP\_SIZE, 1472, 1024, 1024, 512, 128, NUM\_CLASSES]} \\
\hline
6 & \texttt{[INP\_SIZE, 1472, 1024, 1024, 1024, 512, 128, NUM\_CLASSES]} \\
\hline
\end{tabular}
\end{table*}

\begin{table*}[]
\centering
    \caption{Network architecture used for single IPU scaling results with the SHD dataset. Layer sizes for the NMNIST dataset are very similar The specification of \texttt{DENSE\_SIZES} uses python syntax, meaning \texttt{*[x]*[n]} translates to \texttt{n} layers of size \texttt{x}.}
    \label{tab:single_ipu_scaling_shd}
\begin{tabular}{|c|c|r|r|l|}
\hline
\# Neurons per Tile & \# Hidden Layers & Total \# Neurons & \# Weights & \texttt{DENSE\_SIZES} \\
\hline
\hline
2 & 3 & $2,942$ & $2,598,632$ & \texttt{[700, 974, 974, 974, 20]} \\ \hline
4 & 6 & $5,884$ & $5,480,128$ & \texttt{[700, *[980]*2, *[976]*4, 20]} \\ \hline
8 & 12 & $11,764$ & $11,241,504$ & \texttt{[700, *[984]*4, *[976]*8, 20]}  \\ \hline
16 & 24 & $23,524$ & $22,764,736$ & \texttt{[700, *[992]*5, *[976]*19, 20]} \\ \hline
\end{tabular}
\end{table*}

\begin{table*}[]
\centering
    \caption{Network architecture used for single IPU scaling results with the NMNIST dataset. The specification of \texttt{DENSE\_SIZES} uses python syntax, meaning \texttt{*[x]*[n]} translates to \texttt{n} layers of size \texttt{x}. For 16 neurons per tile execution failed due to out of memory error.}
    \label{tab:single_ipu_scaling_nmnist}
\begin{tabular}{|c|c|r|r|l|}
\hline
\# Neurons per Tile & \# Hidden Layers & Total \# Neurons & \# Weights & \texttt{DENSE\_SIZES} \\
\hline
\hline
2 & 3 & $2,942$ & $3,047,428$ & \texttt{[1152, 978, 978, 976, 10]} \\ \hline
4 & 6 & $5,882$ & $5,928,976$ & \texttt{[1152, *[980]*4, *[976]*2, 20]} \\ \hline
8 & 12 & $11,762$ & $11,692,192$ & \texttt{[1152, *[984]*5, *[976]*7, 20]}  \\ \hline
16 & 24 & $23,530$ & $23,234,848$ & \texttt{[1152, *[992]*6, *[976]*20, 20]} \\ \hline
\end{tabular}
\end{table*}

\subsection{Training Convergence and Accuracy}

The specific network architecture sizes used for the training convergence and accuracy results are shown in Table \ref{tab:convergence_network_architecure}. For further hyperparameter and algorithmic choices we refer to the provided code.


\begin{figure*}[t]
\centering
\includegraphics{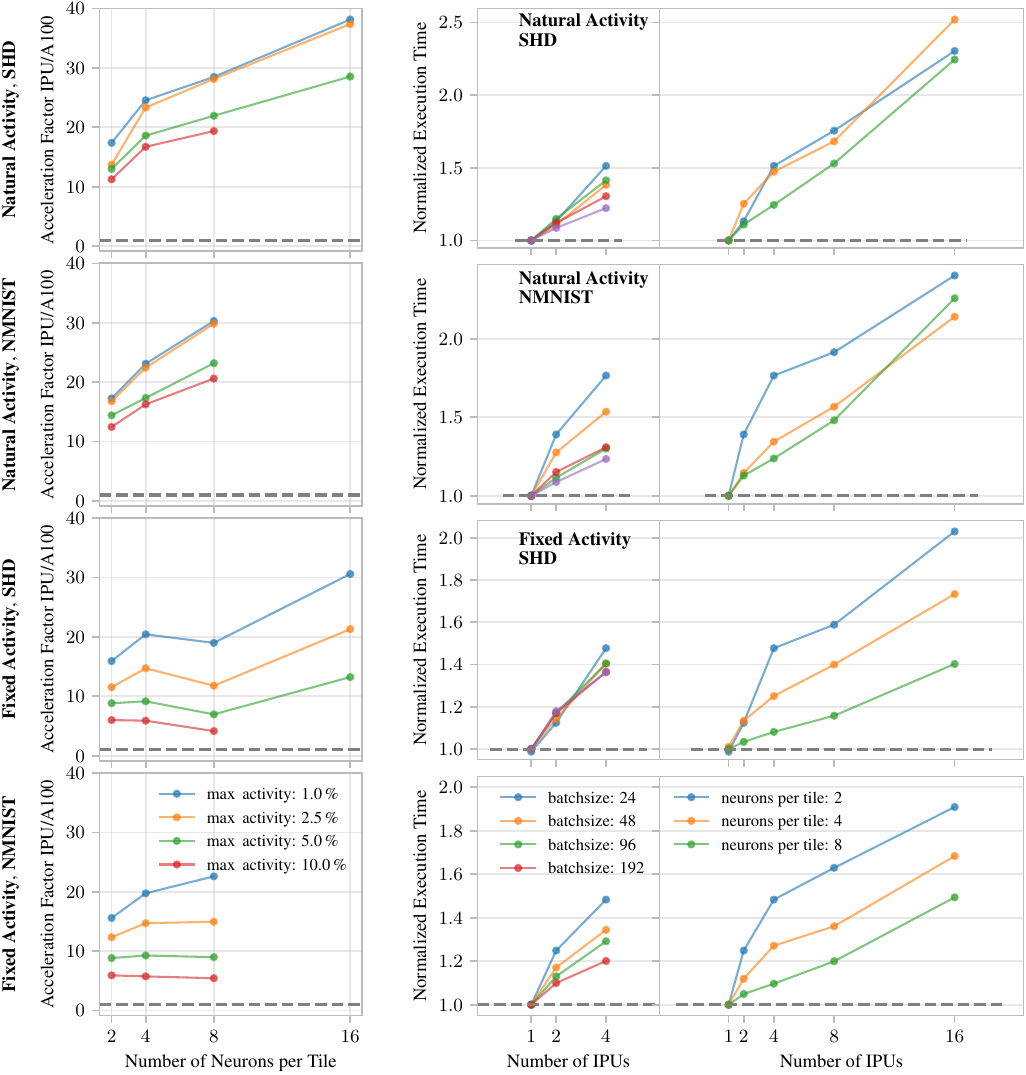}
\caption{Left: Acceleration factor of Sparse IPU vs. A100 implementation for diverse network sizes achieved by adjusting number of neurons per IPU-tile. Right: Weak scaling results at max activity of 5$\%$, exploring compute-workload to communication-latency ratios through batch size variation and network size per IPU adjustment. Top to bottom show results for (1) ``natural activity'' on the SHD datasets, (2) ``natural activity'' on the NMNIST datasets, (3) ``fixed activity'' on the SHD dataset, (4) ``fixed activity'' on the NMNIST dataset.}
\label{fig:results:scaling}
\end{figure*}

\subsection{Scalability Mutli Neuron}

The specific network architectures used for the single IPU multi neuron scalability experiments are shown in Table \ref{tab:single_ipu_scaling_shd} for the SHD dataset and in Table  \ref{tab:single_ipu_scaling_nmnist} for the NMNIST. The scaling to 16 neurons per tile on the NMNIST dataset failed due to out of memory errors. Figure \ref{fig:results:scaling} (left) shows the results for all combinations of ``natural activity'' and ``fixed activity'' with the SHD and the NMNIST dataset. 

Figure \ref{fig:results:scaling} (left) the third row for ``fixed activity'' on the SHD dataset shows a dip in the acceleration factor for the case with 8 neurons per tile, while it is not present in row 1, the equivalent ``natural activity'' case. We attribute this slowdown to a slight flaw in our codebase and not to an inherent issue with the IPU hardware: While we implemented SIMD paralleization for the 2 and 4 neuron per tile setup manually, we relied on the compiler to automatically SIMD parallelize longer loops for more neurons per tile. While the compiler performed as expected for 16 neurons per tile, it has not in the case for 8 neurons per tile, which results in almost identical execution times for the 8 neuron and 16 neuron per tile cases for the compute bound ``fixed activity'' case.

\subsection{Scalability Mutli IPU}

Figure \ref{fig:results:scaling} (right) shows that scaling from 1 to 2 and from 2 to 4 IPUs shows a larger slowdown compared to the additional slowdown incurred by scaling from 4 to 8 or 8 to 16 IPUs. This is expected as intra-IPU communication is much faster than inter-IPU communication via IPU links, resulting in a slowdown when scaling from 1 to 2 IPUs. Additionally, the IPU connectivity graph features higher bandwidth interconnects also between a pair of IPUs which explain the slowdown when scaling from 2 to 4 IPUs. Beyond that, the connectivity between IPUs is handled via the same interconnects which explains the reduced slowdown when scaling beyond 4 IPUs.

\subsection{Execution Traces}

\begin{figure*}[t]
\centering
\includegraphics[width=\linewidth]{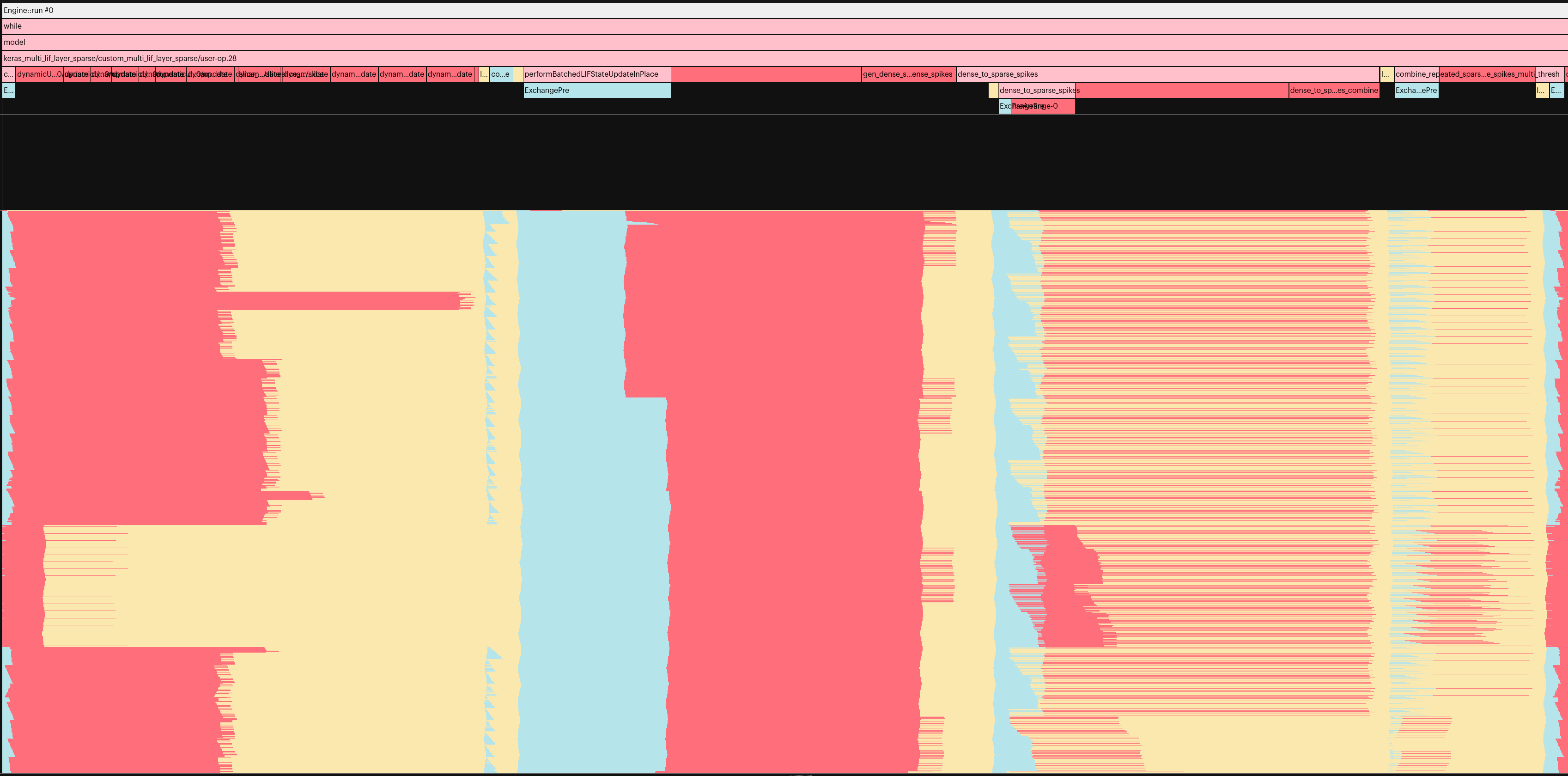}
\includegraphics[width=\linewidth]{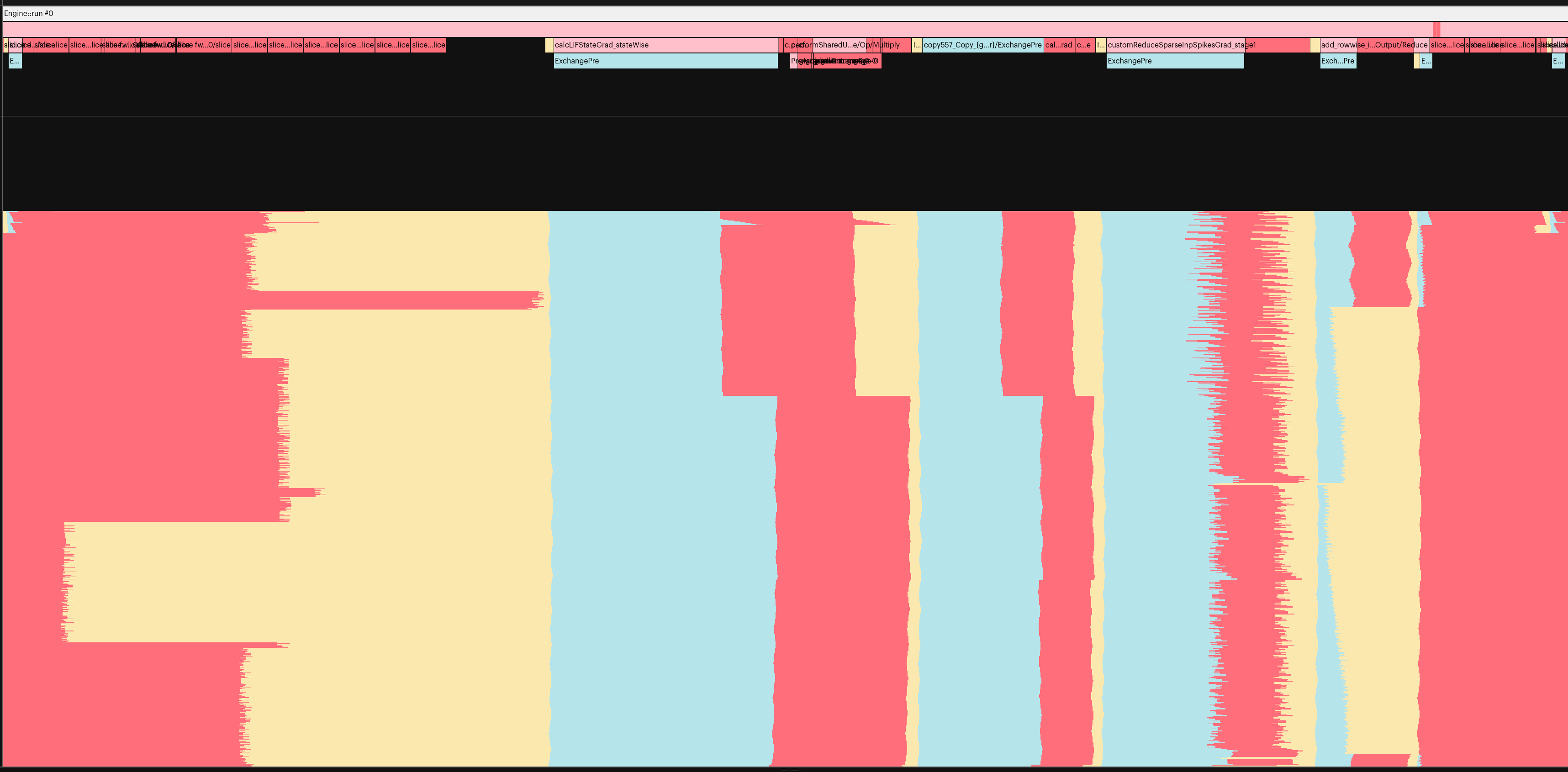}
\caption{Execution traces of the processing of a batch of data of a single algorithmic (SNN) timestep for ``natural activity'' obtained via the popvision tool. The x-axis is the cycle of the program, the y axis depicts the state of a specific tile. The different colors translate to the program state, where blue stands for inter-tile communication within one IPU, red stand for computation phase and yellow for synchronization phase between tiles. Top: Forward pass, Bottom: backward pass.}
\label{fig:execution_natural_activity}
\end{figure*}

\begin{figure*}[t]
\centering
\includegraphics[width=\linewidth]{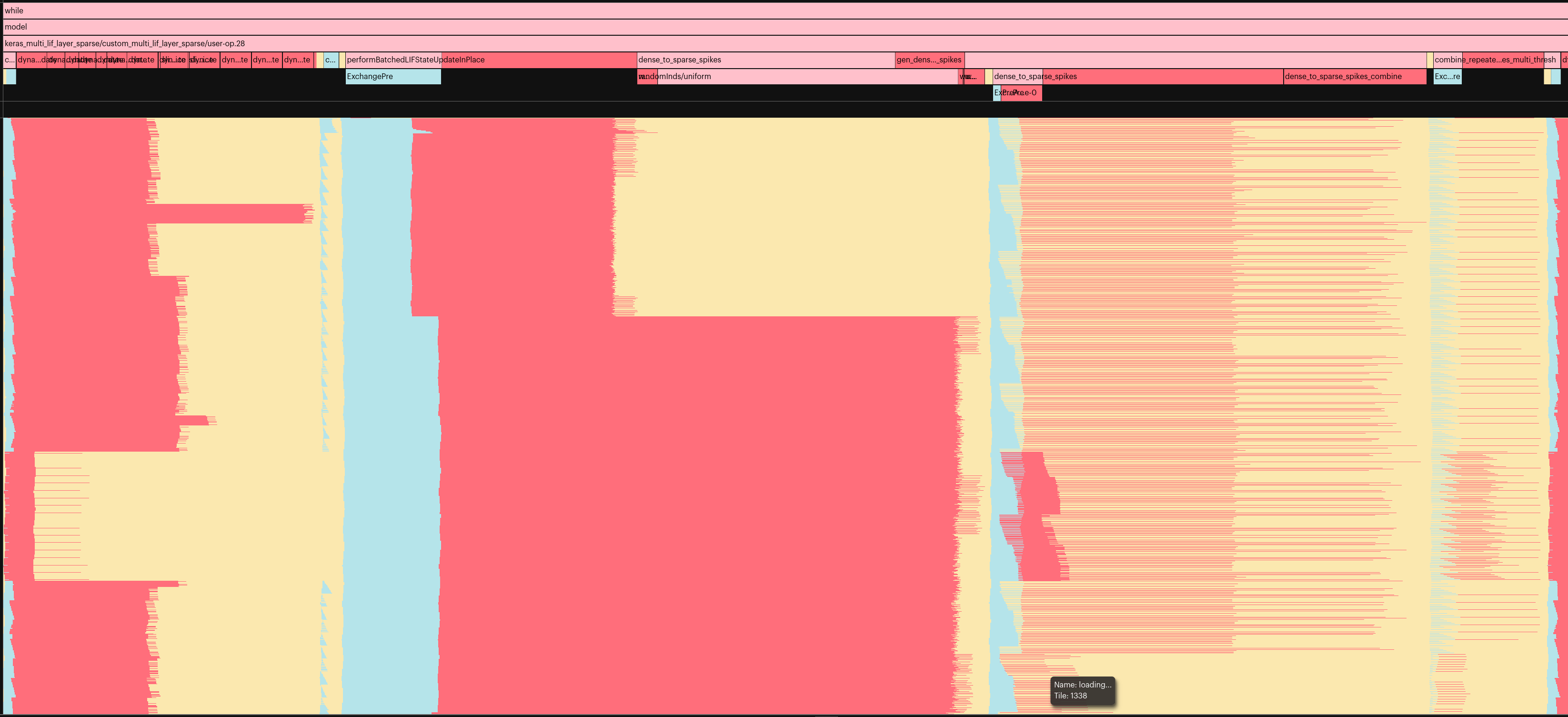}
\includegraphics[width=\linewidth]{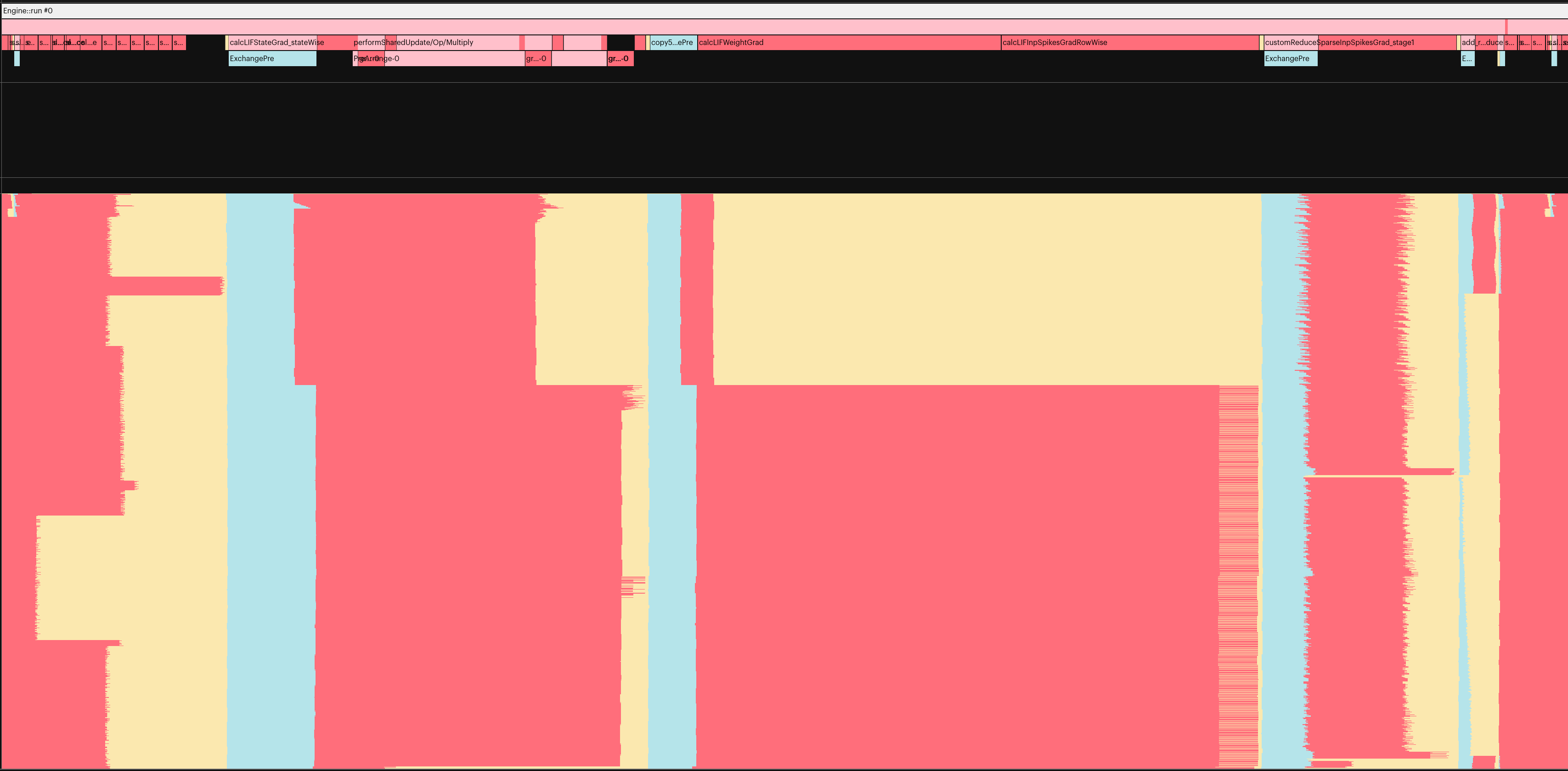}
\caption{Execution traces of the processing of a batch of data of a single algorithmic (SNN) timestep for ``fixed activity'' obtained via the popvision tool. The x-axis is the cycle of the program, the y axis depicts the state of a specific tile. The different colors translate to the program state, where blue stands for inter-tile communication within one IPU, red stand for computation phase and yellow for synchronization phase between tiles. Top: Forward pass, Bottom: backward pass.}
\label{fig:execution_fixed_activity}
\end{figure*}

Figures \ref{fig:execution_natural_activity} and \ref{fig:execution_fixed_activity} show the execution traces for the ``natural activity'' and ``fixed activity'' cases respectively. It becomes clear that for ``fixed activity'' computation is the more dominant part, whereas for ``natural activity'' inter-tile communication within one IPU takes a much larger share of the total time. Additionally, it becomes clear, that further improvements in regard to load balancing are possible that then improve overall execution speed.
For multi-IPU setting an additional inter-IPU communication step takes place not shown in the Figures here.

\end{document}